\documentclass[letterpaper]{article}

\pdfoutput=1

\usepackage{aaai19}
\usepackage{times}
\usepackage{helvet}
\usepackage{courier}
\usepackage{epsfig}
\usepackage{graphicx}
\usepackage{amsmath}
\usepackage{amssymb}
\usepackage{stmaryrd}
\usepackage{eucal}
\usepackage{bm}
\usepackage{subfigure}
\usepackage{caption}
\usepackage{xspace}
\usepackage{enumerate}
\usepackage{multirow}
\usepackage{xcolor}

\begin{document}
% The file aaai.sty is the style file for AAAI Press 
% proceedings, working notes, and technical reports.
\title{Compressing Recurrent Neural Networks with Tensor Ring for Action Recognition}
\author{Yu Pan$^1$, Jing Xu$^1$, Maolin Wang$^1$, Jinmian Ye$^1$, Fei Wang$^2$, Kun Bai$^3$, Zenglin Xu$^1$\footnote{Corresponding Author.}
\\
$^1$University of Electronic Science and Technology of China, Sichuan, China\\
Emails:\{ypyupan, xujing.may, morin.w98, jinmian.y, zenglin\}@gmail.com\\
$^2$Weill Cornell Medical College, Cornell  University, New York, NY, USA\\
Email:few201@cornell.edu\\
$^3$Mobile Internet Group, Tencent Inc., Shenzhen, Guangdong, China\\
Email: kunbai@tencent.com
}

\maketitle
\begin{abstract}
\begin{quote}
 Recurrent Neural Networks (RNNs) and their variants, such as Long-Short Term Memory (LSTM) networks, and Gated Recurrent Unit (GRU) networks, have achieved promising performance in sequential data modeling. The hidden layers in RNNs can be regarded as the memory units, which are helpful in storing information in sequential contexts. However, when dealing with high dimensional input data, such as video and text, the input-to-hidden linear transformation in RNNs brings high memory usage and huge computational cost. This makes the training of RNNs very difficult. To address this challenge,  we propose a novel compact LSTM model, named as TR-LSTM, by utilizing the low-rank tensor ring decomposition (TRD) to reformulate the input-to-hidden transformation. Compared with other tensor decomposition methods, TR-LSTM is more stable. In addition, TR-LSTM can complete an end-to-end training and also provide a fundamental building block for RNNs in handling large input data. Experiments on real-world action recognition datasets have demonstrated the promising performance of the proposed TR-LSTM compared with the tensor-train LSTM and other state-of-the-art competitors.  %outperforming than other low-rank methods and achieving better performance with less computational cost.
% By decomposing the input-to-hidden matrix into the form of Tensor Ring Decomposition(TRD), the TR-RNN model can achieve better performance with less computation in dealing with high-dimensional input data.
%Compared with other methods based on using various feature extractors, the TR-RNN can complete the end-to-end training and provide a fundamental block for RNNs in handling large input data. 
%Compared with the standard LSTM, the TR-LSTM can achieve better recognition accuracy with fewer parameters. For example, TR-LSTM compressed the input-to-hidden layer over 9362 times while with an accuracy improvement over 20.9\%  on the UCF11 dataset. % achieving the best performance among the state-of-the-art RNN training methods based on low-rank tensor decomposition. 
%We also achieve  ~~~~~  accuracy improvement with the compression rate of ~~~~ with feature extractor in UCF11 by comparing to vinilla LSTM.
\end{quote}
\end{abstract}

% the ap \ref{complexity-ana} show
\section{Introduction}
%1. RNNs
Recurrent Neural Networks (RNNs) have achieved great success in analyzing sequential data in various applications, such as computer vision \cite{byeon2015scene,liang2016semantic,theis2015generative}, natural language processing, etc.. Thanks to the ability in capturing long-range dependencies from input sequences~\cite{rumelhart1988learning,sutskever2014sequence}. 
To address the gradient vanishing issue which often leads to the failure of long-term memory in vanilla RNNs, advanced variants such as Gate Recurrent Unit (GRU) 
% \cite{cho2014learning} 
and Long-Short Term Memory (LSTM)
% ~\cite{hochreiter1997long}
have been proposed and applied in many learning tasks~\cite{byeon2015scene,liang2016semantic,theis2015generative}. %They use gate mechanism to enrich the expression of the vanilla RNN. 

% 2.  problems of RNNs, especially in action recognition, 
Despite the success, LSTMs and GRUs suffer from the huge number of parameters, which makes the training process notoriously difficult and easily over-fitting. In particular, in the task of action recognition from videos, a video frame usually forms a high-dimensional input, which makes the size of the input-to-hidden matrix extremely large.  %Specifically, we consider the giant matrix-vector multip
%In the field of the computer vision, there are also many datasets with continuous characteristics, e.g., the high correlation between frames of video. But when we implement plain RNN in CV, it's difficult to handle the high-dimensional inputs data. 
For example, in the UCF11~\cite{DBLP:conf/cvpr/LiuLS09}, a video action recognition dataset, an RGB video clip is a frame with a size of $160\times 120 \times 3$, and  the dimension of the input vector fed to the vanilla RNN can be over 57,000. Assume that the length of hidden layer vector is 256. Then, the input-to-hidden layer matrix has a load of parameters up to 14 millions.  %causing by the long input vector, resulting in high memory usage and huge computational complexity, which makes RNNs unscalable, susceptible to over-fitting and notoriously hard to train.
Although a pre-processing feature extraction step via deep convolutional neural networks can be utilized to obtain static feature maps as inputs to RNNs \cite{DBLP:conf/cvpr/DonahueHGRVDS15}, the over-parametric problem is still not fully solved. %For example, if the extracted feature size is  $I = 4096$ and the number of hidden states is $J=256$, the total number of parameters in calculating the input-to-hidden matrix is $1.0 \times 10^6$. % Moreover, the pretrained CNN extractor on images may neglect the correlation between video frames, thus leading to suboptimal performance.%[This claim supported by \cite{DBLP:conf/icml/YangKT17} but not supported by our experiments.

A promising direction to reduce the parameter size is to explore the low-rank structures in the weight matrices. Inspired from the success of tensor decomposition methods in CNNs~\cite{DBLP:conf/nips/NovikovPOV15,liyycyx17tensor}, various tensor decomposition methods have been explored in RNNs~\cite{DBLP:conf/icml/YangKT17,Ye_2018_CVPR}. In particular, in \cite{DBLP:conf/icml/YangKT17}, the tensor train (TT) decomposition has been applied to RNNs in an end-to-end way to replace the input-to-hidden matrix, and achieved state-of-the-art performance in finding the low-rank structure in RNNs. However, the restricted setting on the ranks and the restrained order of core tensors makes TT-RNN models sensitive to parameter selection. In detail, the optimal setting of TT-ranks is that they are small in the border cores and large in middle cores, e.g., like an olive \cite{DBLP:journals/corr/ZhaoZXZC16}. %Another promising tensor decomposition method is the Block-Term LSTM \cite{Ye_2018_CVPR}. 

To address this issue, we propose to use the tensor ring decomposition (TRD) \cite{DBLP:journals/corr/ZhaoZXZC16} to extract the low-rank structure of the input-to-hidden matrix in RNNs. %In detail, we introduce two forms of tensor ring decomposition: tenor ring in the form of Matrix Product State(MPS), and tensor ring decomposition in the form of Matrix Product Operation~\cite{DBLP:journals/ftml/CichockiLOPZM16}, whose description will be given in the following section. We name these two decomposition ways as TR-MPS and TR-MPO, respectively. 
%It has been demonstrated that the tenor ring decomposition can alleviate the strict constraints in tensor train decomposition via interconnecting the first and the last core tensors circularly, endowed with more powerful expressive ability\cite{zhao2017learning}. Empirical evaluations have shown the superiority of using TRD (in the form of MPS) to replace the fully-connected layers in CNN~\cite{Wang_2018_CVPR}.
Specifically, the input-to-hidden matrices are reshaped into a high-dimensional tensor and then factorized using TRD. Since this corresponding tensor ring layer automatically models the inter-parameter correlations, the number of parameters can be much smaller than the original size of the linear projection layer in standard RNNs. In this way, we present a new TR-RNN model with a similar representation power but with several orders of fewer parameters. In addition, since TRD can alleviate the strict constraints in tensor train decomposition via interconnecting the first and the last core tensors circularly~\cite{DBLP:journals/corr/ZhaoZXZC16}, we expect TR-RNNs to have more expressive power. %Empirical evaluations have shown the superiority of using TRD over tensor train in replacing the fully-connected layers in CNN~\cite{Wang_2018_CVPR}. %Therefore, we expect the proposed TR-RNNs can achieve robust performance over TT-RNNs and standard RNNs.
It is important to note that the tensor ring layer can be optimized in an end-to-end training, and can also be utilized as a building block into current LSTM variants. 
For illustration, we implement an LSTM with the tensor ring layer, named as TR-LSTM.\footnote{Note that the tensor ring layer can also be plugged in the vanilla RNN and GRU. }

We have conducted empirical evaluations on two real-world action recognition datasets, i.e., UCF11 and HMDB51~\cite{DBLP:conf/iccv/KuehneJGPS11}. For a fair comparison with standard LSTM and  TT-LSTM~\cite{DBLP:conf/icml/YangKT17} and BT-LSTM~\cite{Ye_2018_CVPR}, we conduct experiments in an end-to-end training. As shown in Figure \ref{fig:with-frames}, the proposed TR-LSTM has obtained an accuracy value of 0.869, which outperforms the results of standard LSTM (i.e., 0.681), TT-LSTM (i.e.,0.803), and BT-LSTM (i.e., 0.856). Meanwhile, the compression ratio over the standard LSTM is over 34,000, which is also much higher than the compression ratio given by TT-LSTM and BT-LSTM. Moreover, with the output by pre-trained CNN as the input to LSTMs, TR-LSTM has outperformed most previous competitors including LSTM, TT-LSTM, BT-LSTM, and others recently proposed action recognition methods. Since the TR layer can be used as a building block to other LSTM-based approaches, such as the two-stream LSTM~\cite{DBLP:conf/wacv/GammulleDSF17}, we believe the proposed TR decomposition can be a promising approach for action recognition, by considering the tricks used in other state-of-the-art methods.

\section{Model}
To handle the high dimensional input of RNNs, we introduce the tensor ring decomposition to represent the input-to-hidden layer in RNNs in a compact structure. In the following, we first present preliminaries and background of tensor decomposition, including graphical illustrations of the tensor train decomposition and the tensor ring decomposition, followed by our proposed LSTM model, namely, TR-LSTM.
 
 %.  the details of TTD and TRD, analyze the excellent performance of TRD is because that it's a linear combination of TTs. Then, We introduce TRD in RNNs to get our TR-LSTM model and analyze the compression ratio.
%To fix the over-parametric weight matrix in RNNs from input-to-hidden layer, replacing the parameters in RNNs with the low-rank tensor structure can compress the networks with fewer orders of parameters, and improve the generalization ability without losing representation power. Our model uses TR to replace the weight matrix from the input-to-hidden layer in the RNNs. The new TR-RNN models are compressed under the condition of improved precision, especially suitable for large-scale data inputs in the field of CV.

%Our model is elaborated in this section. First, we introduce preliminaries and background of tensor decomposition, describe the details of TTD and TRD, analyze the excellent performance of TRD is because that it's a linear combination of TTs. Then, We introduce TRD in RNNs to get our TR-LSTM model and analyze the compression ratio.

\subsection{Preliminaries and Background}

\subsubsection{Notation}
\begin{figure}[t]
\centering
\subfigure[a $L_1 \times L_2$ matrix]{
  \label{fig:tensor_notation_a}
  \includegraphics[scale=1.07]{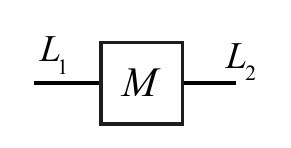}
  }
\subfigure[matrix contraction]{
  \label{fig:tensor_notation_b}
  \includegraphics[scale=1.07]{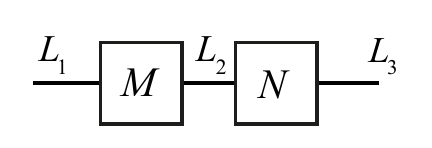}
  }
\subfigure[a $L_1\times L_2\times L_3$ tensor]{
  \label{fig:tensor_notation_c}
  \includegraphics[scale=1.1]{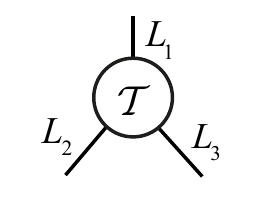}
%   \begin{minipage}
%   \caption{hello}
%   \end{minipage}
  }

\caption{Tensor diagrams. (a) shows the graphical representation of the matrix $\mathbf{M}\in \mathbb{R}^{L_1\times L_2}$, where $L_1$ and $L_2$ denote the matrix size. A matrix is represented by a rectangular node. (b) demonstrates the contraction between two matrices(tensors), which is represented by an axis connecting them together and the contraction between $\mathbf{M}$ and $\mathbf{N}$ resulting in a new matrix with shape $\mathbb{R}^{L_1\times L_3}$. (c) presents the graphical notation of a tensor $\pmb{\mathcal{T}}\in \mathbb{R}^{L_1\times L_2 \times L_3}$.}
\label{fig:tensor_notation}
\end{figure}

In this paper, a $\mathit{d}$-order tensor, e.g., $\pmb{\mathcal{D}}\in\mathbb{R}^{L_1\times L_2 \dots \times L_d} $ is denoted by a boldface Euler script letter. With all subscripts fixed, each element of a tensor is expressed as: $\pmb{\mathcal{D}}_{l_1,l_2,\dots l_d}\in \mathbb{R}$. Given a subset of subscripts, we can get a sub-tensor. For example, given a subset $\{L_1=l_1, L_2=l_2\}$, we can obtain a sub-tensor $\pmb{\mathcal{D}}_{l_1, l_2} \in \mathbb{R}^{L_3 \dots \times L_d}$. Specifically, we denote a vector by a bold lowercase letter, e.g., $\mathbf{v} \in \mathbb{R}^L$, and matrices by bold uppercase letters, e.g., $\mathbf{M} \in \mathbb{R}^{L_1 \times L_2}$. We regard vectors and matrices as 1-order tensors and 2-order tensors, respectively. Figure~\ref{fig:tensor_notation} draws
the tensor diagrams presenting the graphical notations and the essential operations.

\subsubsection{Tensor Contraction}
Tensor contraction can be performed between two tensors if some of their dimensions are matched. For example, given two  3-order tensors $\pmb{\mathcal{A}}\in\mathbb{R}^{L_1\times L_2 \times L_3} $ and $\pmb{\mathcal{B}}\in\mathbb{R}^{J_1\times J_2 \times J_3}$, when $L_3 = J_1$, the contraction between these two tensors result in a tensor with the size of  $L_1\times L_2 \times J_2 \times J_3$, where the matching dimension is reduced, as shown in  Equation~(\ref{eq:tensor_contraction}):
\begin{equation}
    (\pmb{\mathcal{A}}\pmb{\mathcal{B}})_{l_1,l_2,j_2,j_3} = \pmb{\mathcal{A}}_{l_1,l_2}\pmb{\mathcal{B}}_{j_2,j_3}
    = \sum_{p=1}^{L_3} \pmb{\mathcal{A}}_{l_1,l_2,p }\pmb{\mathcal{B}}_{p,j_2,j_3}
\label{eq:tensor_contraction}
\end{equation}

\begin{figure}
\centering
	\includegraphics[]{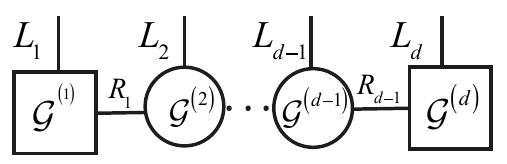}
\caption{Tensor Train Decomposition: Noted that in tensor train the rank $R_0$ and $R_d$ are constrained to 1, so the first and the last core are matrices while the inner cores are 3-order tensors.}
\label{Fig:tensor_train_decomposition}
\end{figure}

\subsubsection{Tensor Train Decomposition}
Through the tensor train decomposition (TTD), a high-order tensor can be decomposed as the product of a sequence of low-order tensors. For example, a $d$-order tensor $\pmb{\mathcal{D}}\in\mathbb{R}^{L_1\times L_2 \dots \times L_d} $ can be decomposed as follows:
\begin{align}
    \pmb{\mathcal{D}}_{l_1,l_2,\dots, l_d}
    &\overset{TTD}{=}\pmb{\mathcal{G}}^{(1)}_{l_1}\pmb{\mathcal{G}}^{(2)}_{l_2}\dots\pmb{\mathcal{G}}^{(d)}_{l_d} \notag \\ 
    & \overset{\quad \quad}{=} \sum_{r_1,r_2,\dots,r_{d-1}}\pmb{\mathcal{G}}^{(1)}_{r_0,l_1,r_1}\dots \pmb{\mathcal{G}}^{(d)}_{r_{d-1},l_d,r_d}
     \label{eq:TT_decom}
\end{align}
where each  $\pmb{\mathcal{G}}^{(k)}\in\mathbb{R}^{R_{k-1}\times L_k \times R_k}$ is called a core tensor. The tensor train rank, shorted as TT-rank, $[R_0, R_1, R_2, \dots, R_d]$, for each $k \in \{0, 1, \dots , d\}$, $0 < r_k \leq R_k $, corresponds to the complexity of tensor train decomposition.  Generally, in tensor train decomposition, the constraint $R_0 = R_d = 1$ should be satisfied and other ranks are chosen manually. Figure~\ref{Fig:tensor_train_decomposition} illustrates the form of tensor train decomposition.

\begin{figure}[t]
\centering
\subfigure[TRD in a ring form ]{
\label{fig:TR_TRD in TR form}
   \includegraphics[]{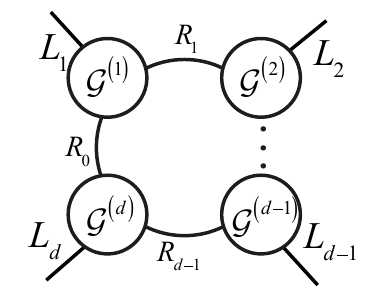}}
\centering
\subfigure[TRD as the sum of TTs ($r_0=r_d$)]{
\label{fig:TRD in the sum of TTs form}
 \includegraphics[]{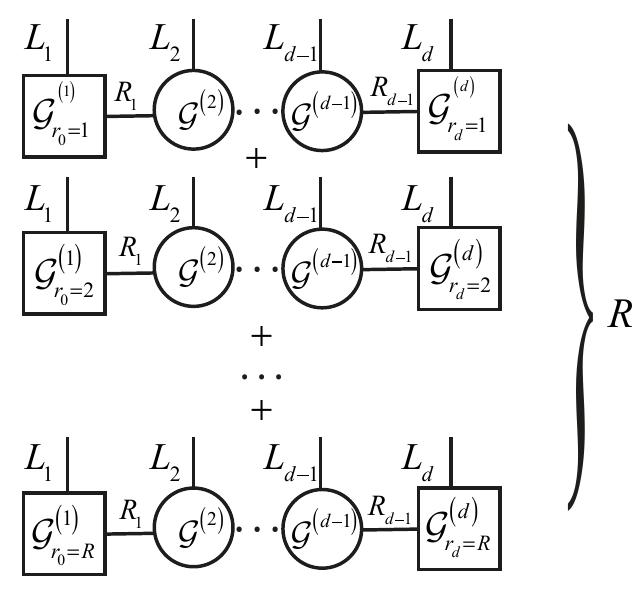}}
\caption{Two representations of tensor ring decomposition (TRD). In Figure~\ref{fig:TR_TRD in TR form}, TRD is expounded in the traditional way: the core tensors are multiplied one by one, and form a ring structure. In Figure~\ref{fig:TRD in the sum of TTs form}, TRD is illustrated in an alternative way: the summation of a series of tensor trains. By fixing the subscript $r_0 $ of $\pmb{\mathcal{G}}^{(1)}$ and $r_d$ of $\pmb{\mathcal{G}}^{(d)}$: $r_0 = r_d = k$, where $k \in \{1, 2, \dots, R\}$, both $\pmb{\mathcal{G}}^{(1)}$ and $\pmb{\mathcal{G}}^{(d)}$ are divided into $R$ matrices.}
\label{fig:TR}
\end{figure}

\subsubsection{Tensor Ring Decomposition}
The main drawback in tensor train decomposition is the limit of the ranks' setting, which hinders the representation ability and the flexibility of the TT-based models. At the same time, a strict order must be followed when multiplying TT cores, so that the alignment of the tensor dimensions is extremely important in obtaining the optimized TT cores, but it is still a challenging issue in finding the best alignment\cite{DBLP:journals/corr/ZhaoZXZC16}.

In the tensor ring decomposition(TRD), an important modification is interconnecting the first and the last core tensors circularly and constructing a ring-like structure to alleviate the aforementioned limitations of the tensor train. Formally, we set $R_0 = R_d = R$ and $R\geq 1$, and conduct the decomposition as:
\begin{align}
    \pmb{\mathcal{D}}_{l_1,l_2,\dots, l_d} 
    \overset{TRD}{=}&\sum_{r_0=r_d, r_2, \dots, r_{d-1}}\notag \\
    &\pmb{\mathcal{G}}^{(1)}_{r_0,l_1, r_1}\pmb{\mathcal{G}}^{(2)}_{r_1, l_2, r_2}\dots\pmb{\mathcal{G}}^{(d)}_{r_{d-1}, l_d,r_d}
    \label{eq:TR_decom}
\end{align}

For a $d$-order tensor, by fixing the index $k$ where  $k \in \{1, 2, \dots, R\}$, the first order of the beginning core tensor $\pmb{\mathcal{G}}^{(1)}_{r_0 = k}$ and the last order of the ending core tensor $\pmb{\mathcal{G}}^{(d)}_{r_d = k}$,  can be reduced to  matrices. Thus, along each of the $R$ slices of $\pmb{\mathcal{G}}^{(1)}$, we can separate the tensor ring structure as a summation of $R$ of tensor trains. For example, by fixing $r_0=r_d=k$, the  product of $\pmb{\mathcal{G}}^{(1)}_{k,l_1}\pmb{\mathcal{G}}^{(2)}_{l_2}\dots\pmb{\mathcal{G}}^{(d)}_{l_d,k}$ has the form tensor train decomposition. Therefore, the tensor ring model is essentially the linear combination of $R$ different tensor train models. Figure \ref{fig:TR} demonstrates the tensor ring structure, and the alternative interpretation as a summation of multiple tensor train structures.

\subsection{TR-RNN model}
The core conception of our model is elaborated in this section. By transforming the input-to-hidden weight matrices in TR form, and applying them into RNN and its variants, we get our TR-RNN models. 
% TRD is used to reconstruct the input-to-hidden matrix multiplication $\mathbf{y} = \mathbf{W}\mathbf{x}$ where $\mathbf{W}\in \mathbb{R}^{I\times O}$. $\mathbf{x}\in \mathbb{R}^{I}$ is the input vector, and $\mathbf{y}\in \mathbb{R}^{O}$ is the output vector. We first reshape the input vector $\mathbf{x}$ and weight matrix $\mathbf{W}$ into higher dimension tensors, then transform the weight tensor into a TR form. Finally, the tensor contraction between the input and weight tensor will be conducted to get the output tensor. 

\subsubsection{Tensorizing $\mathbf{x}$, $\mathbf{y}$ and $\mathbf{W}$}
Without loss of generality, we tensorize the input vector $\mathbf{x} \in \mathbb{R}^I$, output vector $\mathbf{y}\in\mathbb{R}^O$, and weight matrix $\mathbf{W} \in \mathbb{R}^{I\times O}$ into tensors $\pmb{\mathcal{X}}$, $\pmb{\mathcal{Y}}$, and $\pmb{\mathcal{W}}$, shown in Equation~(\ref{eq:w_x_y_shape}):
\begin{align}
    \pmb{\mathcal{X}}\in \mathbb{R}^{I_1\times I_2\times ,\dots, \times I_n} , \pmb{\mathcal{Y}}\in \mathbb{R}^{O_1\times O_2\times ,\dots, \times O_m}  \notag \\
    \pmb{\mathcal{W}}\in \mathbb{R}^{I_1\times I_2\times ,\dots, \times I_n\times O_1\times O_2 ,\dots, \times O_{m}} 
\label{eq:w_x_y_shape}
\end{align}
where
$$\prod_{i=1}^{n}I_i = I, ~~~\prod_{j=1}^{m}O_j = O $$

\begin{figure}
\centering
	\includegraphics[]{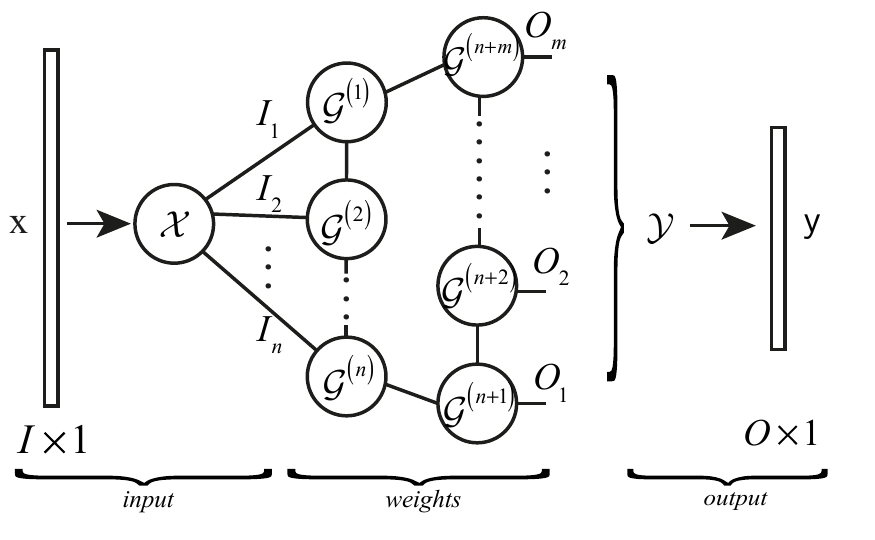}
\caption{TRL: $\pmb{\mathcal{X}}$ represents the input tensor with shape $\mathbb{R}^{I_1\times I_2\times ,\dots, \times I_n}$ after reshaping the input vector $\mathbf{x}\in\mathbb{R}^{I\times 1}$. By performing the multiplication operation shown in Equation~(\ref{eq:mps_net}) with the weights in TRD form, the output tensor $\pmb{\mathcal{Y}}$ with shape $\mathbb{R}^{O_1\times O_2\times ,\dots, \times O_m}$can be obtained. Then, after transforming $\pmb{\mathcal{Y}}$ into vector, we can get the final output vector $\mathbf{y}\in\mathbb{R}^{O}$.}
\label{Fig:TR_MPS_Layer}
\end{figure}

\subsubsection{Decomposing $\mathbf{W}$}
For an $n$-order input and $m$-order output, we decompose the weight tensor into the form of TRD with $n+m$ core tensors multiplied one by one, each of which is corresponding to an input dimension or an output dimension, referring to Equation~(\ref{eq:W_TR_MPS}). Without loss of generality, the core tensors corresponding to the input dimensions and output dimensions are grouped respectively, as shown in Figure~\ref{Fig:TR_MPS_Layer}.
% TR-MPS is another representation of TRD. In this form of decomposition, we have n-dimensional input tensor, m-dimensional output tensor, $(n+m)$-dimensional weight tensor. We decomposite the $(n+m)$ dimension weight tensor using the representation of the matrix product state in Equation~(\ref{eq:W_TR_MPS}). After decomposing weight tensor, each core tensor has either an input dimension or an output dimension. So, there exists $(n+m)$ core tensors in TR-MPS. N of them are input core tensors and m of them are output core tensors. The dimension of each core tensor reduces from 4 to 3 compared with TR-MPO. In Figure~\ref{Fig:TR_MPS_Layer}, the weights part is TRD in the MPS representation.
\begin{align}
& TRD(\pmb{\mathcal{W}})_{i_1,\dots , i_n,o_1, \dots, o_m}= \sum_{r_0,\dots, r_n, r_{n+1}, \dots, r_{n+m-1}}\notag \\
& \pmb{\mathcal{G}}^{(1)}_{r_0,i_1,r_1} \dots \pmb{\mathcal{G}}^{(n)}_{r_{n-1}, i_n, r_n} \pmb{\mathcal{G}}^{(n+1)}_{r_{n}, o_1, r_{n+1}}\dots \pmb{\mathcal{G}}^{(n+m)}_{r_{n+m-1}, o_{m},r_0}
\label{eq:W_TR_MPS}
\end{align}

The tensor contraction from input to hidden layer in TR form is shown in Equation~(\ref{eq:mps_net}). We multiply  the  input  tensor with input core tensors and output core tensors sequentially. The complexity analysis of forward and backward process is elaborated in the appendix.
\begin{align}
\pmb{\mathcal{Y}}_{o_1, o_2, \dots, o_m}= \sum_{i_1,\dots, i_n} \pmb{\mathcal{X}}_{i_1, \dots, i_n} TRD(\pmb{\mathcal{W}})_{i_1, \dots, i_n,o_1, \dots, o_m} 
\label{eq:mps_net}
\end{align}

Compared with the redundant input-to-hidden weight matrix, the compression ratio in TR form is shown in Equation~(\ref{eq:compression_ratio_TR-MPS}).
\begin{align}
C_{TRD}  = \frac{\prod_{i = 1}^{i=n} I_i\prod_{j = 1}^{ j=m}O_j}{\sum_{i=1}^{n}R_{i-1}I_iR_i + \sum_{j=1}^{m}R_{n+j-1}O_jR_{n+j}}
\label{eq:compression_ratio_TR-MPS}
\end{align}

% \begin{figure}[t]{}{}
% \centering
% \subfigure[]{
%   \label{fig:fig_4_node}
%    \includegraphics[]{figure/fig_4_node.pdf}}

% \centering
% \subfigure[]{
% \label{fig:fig_4_node_shape1}
%  \includegraphics[]{figure/fig_4_node_shape1.pdf}}
% \centering
% \subfigure[]{
% \label{fig:fig_4_node_shape2}
%    \includegraphics[]{figure/fig_4_node_shape2.pdf}}
% \caption{}
% \end{figure}

% \subsection{Compression Ratio}
% We analyze the compression ratio of TR-MPO and TR-MPS compare to
% \subsubsection{Compression Ratio of TR-MPO}
% In TR-MPO-LSTM, We can get the total number of parameters as follows, by reshaping the original input to gate units weight matrix into TR-MPO form using Equation~(\ref{eq:W_TR_MPO}):

% \begin{align}
% P_{TR-MPO} = \sum_{i=1}^{n+m}r_{i-1}I_iO_ir_i
% \end{align}

% where $r_0=r_{n+m}$. The number of parameters in vanilla RNN weight matrix is $P_{RNN} = I\times O = \prod_{i = 1}^{i=n+m} I_iO_i$. So the compression ratio of the input-to-hidden layer matrix is:

% \begin{align}
% r  = \frac{P_{RNN}}{P_{TRD}}
% \end{align}

\subsubsection{Tensor Ring Layer (TRL)}
After reshaping the input vector $\mathbf{x}$ and the weight matrix $\mathbf{W}$ into tensor, and decomposing weight tensor into TR representation, 
% After tensorizing the input vector $\mathbf{x}$ to tensor $\pmb{\mathcal{X}}$, the input-to-hidden layer matrix $\mathbf{W}$ to tensor $\pmb{\mathcal{W}}$ and decomposing tensor $\pmb{\mathcal{W}}$ into TR representation, 
we can get the output tensor $\pmb{\mathcal{Y}}$ by manipulating $\pmb{\mathcal{W}}$ and $\pmb{\mathcal{X}}$. The final output vector $\mathbf{y}$ can be obtained by reshaping the output tensor $\pmb{\mathcal{Y}}$ into vector. Because the weight matrix is factorized with TRD, we denote the whole calculation from the input vector $\mathbf{x}$ to output vector $\mathbf{y}$ as tensor ring layer (TRL):
\begin{align}
\mathbf{y} = TRL(\mathbf W, \mathbf x)
\label{eq:TRL}
\end{align}
which is illustrated in Figure~\ref{Fig:TR_MPS_Layer}.
% The illustration of TRL is showed in Figure~ \ref{Fig:TR_MPS_Layer}.

\subsubsection{TR-RNN}
By replacing the multiplication between weight matrix $\mathbf W_{hx}$ and input vector $\mathbf{x}$ with TRL in vanilla RNN. We get our TR-RNN model. The hidden state at time $t$ can be expressed as:
\begin{align}
\mathbf{h}_t = \sigma (TRL(\mathbf W_{hx}, \mathbf x_t) + \mathbf U_{hh}\mathbf h_{t-1} + \mathbf b)
\end{align}
where $\sigma(\cdot)$ denotes the sigmoid function and the hidden state is denoted by $\mathbf{h}_t$. The input-to-hidden layer weight matrix is denoted by $\mathbf W_{hx}$, and $\mathbf U_{hh}$ denotes the hidden-to-hidden layer matrix.
 % was calculated by the input-to-hidden layer matrix: $\mathbf W_{hx}$, the hidden-to-hidden layer matrix: $\mathbf U_{hh}$, the input at time $t$: $\mathbf x_t$, the hidden layer vector at time $t-1$: $\mathbf h_{t-1}$ and the bias vector: $\mathbf b$.

\subsubsection{TR-LSTM}
By applying TRL to the standard LSTM, which is the state-of-the-art variant of RNN,
we can get the TR-LSTM model as follows.
% The vanilla RNN has some limitations. The biggest one is the gradient vanishing issue, which resulting in lacking of the ability to model long-term dependencies. The state-of-the-art variants of vanilla RNN is LSTM,  which uses gating units to decide what to keep in (and what to forget from) memory. Our method can be easily extended to LSTM, by implementing TRL on the input vector to gating units vectors in LSTM:
\begin{align}
&\mathbf{k}_t = \sigma (TRL(\mathbf W_{kx}, \mathbf x_t) + \mathbf U_{kh}\mathbf h_{t-1} + \mathbf b_k) \notag \\
&\mathbf{f}_t = \sigma (TRL(\mathbf W_{fx}, \mathbf x_t) + \mathbf U_{fh}\mathbf h_{t-1} + \mathbf b_f) \notag \\
&\mathbf{o}_t = \sigma (TRL(\mathbf W_{ox}, \mathbf x_t) + \mathbf U_{oh}\mathbf h_{t-1} + \mathbf b_o) \notag \\
&\mathbf{g}_t = \tanh(TRL(\mathbf W_{gx}, \mathbf x_t) + \mathbf U_{gh}\mathbf h_{t-1} + \mathbf b_g) \notag \\
&\mathbf{c}_t = \mathbf{f}_t \odot \mathbf{c}_{t-1} + \mathbf{k}_t  \odot \mathbf{g}_t \notag \\
&\mathbf{h}_t = \mathbf{o}_t \odot \tanh(\mathbf{c}_t),
\label{eq:TR_LSTM}
\end{align}
where $\odot$, $\sigma(\cdot)$ and $\tanh(\cdot)$ denote the element-wise product, the sigmoid function and the hyperbolic function, respectively. The weight matrices $\mathbf W_{*x}$ (where $*$ can be $k,f,o,$ or $g$) denote the mapping from the input to hidden matrix, for the input gate  $\mathbf{k}_t$, the forget gate $\mathbf{f}_t$, the output gate $\mathbf{o}_t$,  and the cell update vector $\mathbf{c}_t$, respectively.  The weight matrice $\mathbf U_{*h}$ are  defined similarly for the hidden state $\mathbf h_{t-1}$.

% \subsubsection{Computational Complexity Analysis}
\textbf{Remark.}
 As shown in Equation~(\ref{eq:mps_net}) and demonstrated in Figure~\ref{Fig:TR_MPS_Layer},  the multiplication between the input tensor data $\mathcal{X}$ and the input core tensors $\mathcal{G}^{(i)}$ (for $i=1,\ldots,n$) will produce a hidden matrix in the size of $R_0\times R_n$. It is important to note that the size of the hidden matrix is much smaller than the original data size. In some sense, the ``compressed" hidden matrix can be regarded as the information bottleneck~\cite{{DBLP:journals/corr/physics-0004057},{DBLP:journals/corr/Shwartz-ZivT17}}, which seeks to achieve the balance between maximally compressing the input information and preserving the prediction information of the output. Thus the proposed TR-LSTM has high potentials to reduce the redundant information in the high-dimensional input while achieving good performance compared with the standard LSTM. 
 
 %Compared with the standard LSTM, we can change the size of the matrix manually by choosing the hyper-parameters $R_0$ and $R_n$. Meanwhile, the low rank structure of TR which can reduce the redundant information in input may make the matrix contain more valid information for the output prediction. At the same time, the input is n-order tensor in our model, compared with 1-order vector in vanilla LSTM, the high-dimensional information can be extracted. These may be the reasons why our model is superior to the standard LSTM in dealing with high-dimensional input data.

%The reason that TR-LSTM outperforms the vanilla LSTM may be explained from the perspective of Information Bottleneck, where a supervised learning task can be viewed as a compression problem which seeks to achieve the balance between maximally compressing the input information and preserving the prediction information of the output~\cite{DBLP:journals/corr/physics-0004057}. Achieving fewer parameters including more information is the reason which leads to good results.
\section{Experiments}

To evaluate the  proposed TR-LSTM model, we first design a synthetic  experiment to validate the advantage of tensor ring decomposition over the tensor train decomposition. Through two real-world action recognition datasets, i.e.,  UCF11(YouTube action dataset)~\cite{DBLP:conf/cvpr/LiuLS09} and HMDB51~\cite{DBLP:conf/iccv/KuehneJGPS11}, we evaluate our model from two settings: (1) end-to-end training, where video frames are directly fed into the TR-LSTM; and (2) pre-training to obtain features prior to LSTMs, where a pre-trained CNN was used to extract meaningful low-dimensional features and then forwarded these features to the TR-LSTM. For a fair comparison, we first compare our proposed method with the standard LSTM and previous low-rank decomposition methods, and then with the state-of-the-art action recognition methods.

%\subsection{Synthetic Experiment}
%To verify the effectiveness of tensor decomposition methods in recover the original weights, we design a synthetic dataset. Firstly, we generate input data $x$ from a Gaussian distribution ${{N}(0, 0.5)}$. Then calculate $y$ with a distribution of $\mathbf{y} = \mathbf{W}\mathbf{x}$. Finally, add a small noise into ${x}$ to build a dataset.

%In this section, we design experiments to evaluate the performance of the proposed TR-LSTM. Following literature, we adopt two large video classificaton datasets: UCF11 (YouTube action dataset)~\cite{DBLP:conf/cvpr/LiuLS09} and HMDB51~\cite{DBLP:conf/iccv/KuehneJGPS11}. For fair comparison, we first compare our proposed method with the standard LSTM and previous low-rank decomposition methods, and then with the state-of-the-art action recognition methods.

\subsection{Synthetic Experiment}
\begin{figure}[h]
\centering
\subfigure[Ground truth W]{
    \label{fig:truth3}
    \includegraphics[width=.18\textwidth]{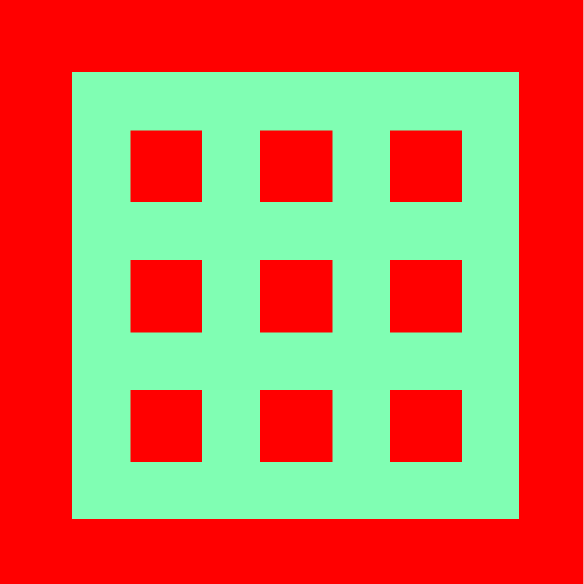}
}
\quad
\subfigure[Linear Regression]{
    \label{fig:line3}
    \includegraphics[width=.18\textwidth]{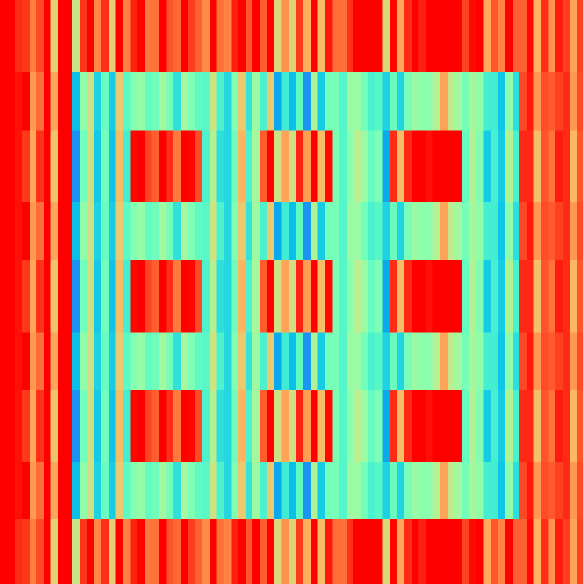}
}

\subfigure[Tensor Train]{
    \centering
    \label{fig:tt3}
    \includegraphics[width=0.18\textwidth]{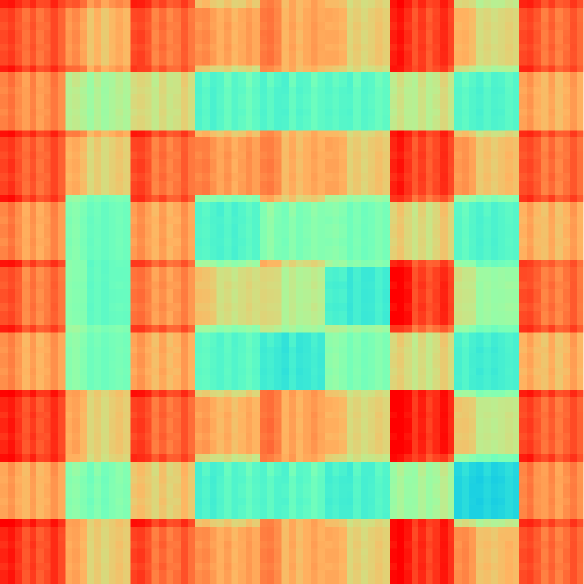}
}
\quad
\subfigure[Tensor Ring]{
    \label{fig:tr3}
    \includegraphics[width=0.18\textwidth]{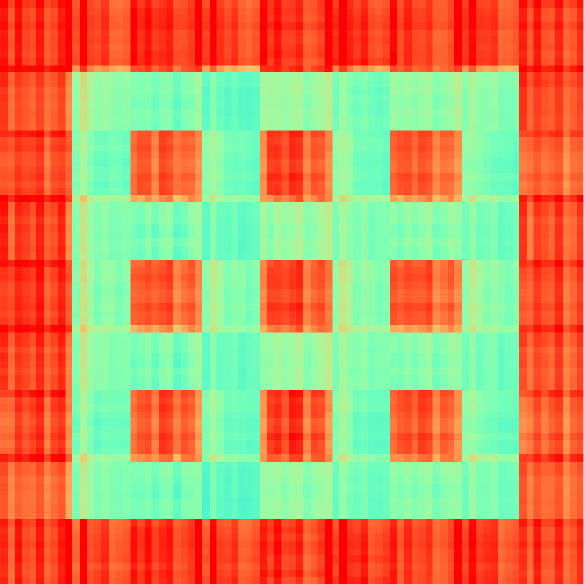}
}
    
\caption{The  illustration on the ground truth $\mathbf{W}$ and the recovered weights from different models. The recovered RMSEs of the linear model, tensor train, and tensor ring, are 0.16, 0.18, and 0.09, respectively. }
\label{fig:model-analysis}
\end{figure}

To verify the effectiveness of tensor decomposition methods in recover the original weights, we design a synthetic dataset. %Firstly, we generate input data $x$ from a Gaussian distribution ${{N}(0, 0.5)}$. Then calculate $y$ with a distribution of $\mathbf{y} = \mathbf{W}\mathbf{x}$. Finally, add a small noise into ${x}$ to build a dataset.
%To demonstrate the superior performance of TR model, we construct a simple controlled experiment to evaluate the quality of the recovered weight matrix. 
Given a low-rank weight matrix $\mathbf{W} \in \mathbb{R}^{81 \times 81}$, which is illustrated in Figure~\ref{fig:truth3}. We first sample 3200 examples, and each dimension follows a normal distribution, i.e.,  $\mathbf{x} \sim \mathcal{N}(0, 0.5\mathbf{I})$ where $\mathbf{I}\in \mathbb{R}^{81}$ is the identity matrix. We then calculate $\mathbf{y}$ according to $\mathbf{y} =  \mathbf{Wx} + \epsilon$ for each $\mathbf{x}$ where $\epsilon \sim \mathcal{N}(0, {\sigma}^2\mathbf{I})$ is a random Gaussian noise and ${\sigma}^2$ is the variance. 
Since the $\mathbf{y}$ is generated from $\mathbf{x}$, the recovered weight matrix should be similar to the ground truth. We use the root mean square error (RMSE) to measure the performance. Should be noted that since the purpose of this experiment is to provide a qualitative and intuitive comparison, we do not add any regularization to the models.

%we build a regression model with various $\mathbf{W}$ substitutions, for the vanilla linear regression, TT model and TR model respectively. To avoid over-fitting and to evaluate the tolerance of noise, we inject various Gaussian Noise $\mathcal{N}(0, \sigma)$ to the input data $\mathbf{x}$. 

\begin{figure}[h]
\centering
\includegraphics[scale=0.3]{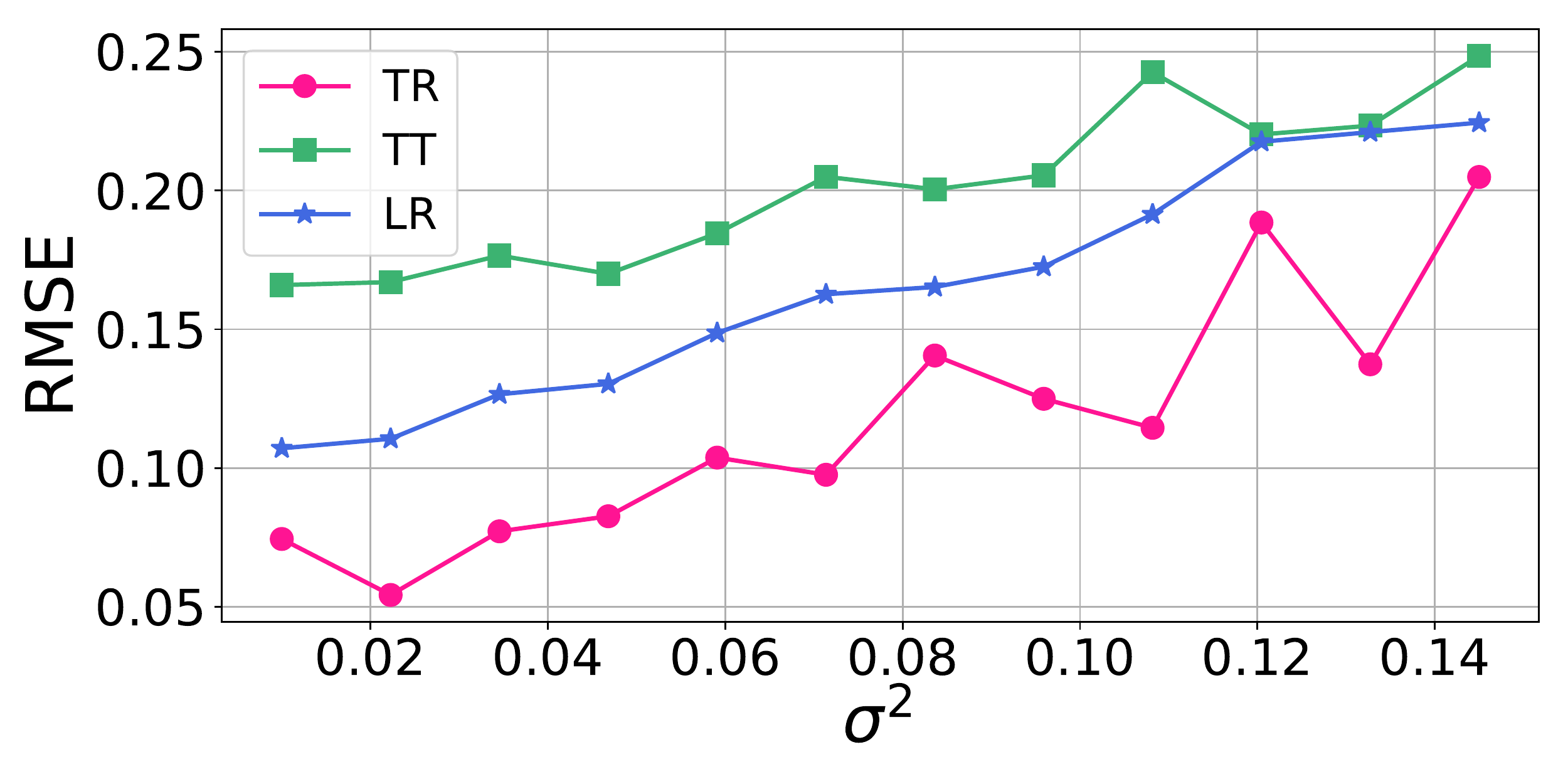}
\caption{The illustration on how the RMSEs of the linear regression(LR), the tensor ring (TR)  and the tensor train (TT) change with added noises.}
\label{fig:sigma3}
\end{figure}

Based on the input data and responses, we estimate the weight matrix $\mathbf{W}$ by running the linear regression, tensor train decomposition, and tensor ring decomposition, respectively. For tensor train and tensor ring, we first reshape input data to a tensor of ${3 \times 3 \times 3 \times 3}$, and reshape the weight matrix to a tensor of the same size. For illustration, Figure~\ref{fig:model-analysis} shows one of the recovered $\mathbf{W}$ (reshaped as a matrix) for the three models when the noise variance is set to 0.05 . Clearly, the proposed tensor ring model performs the best among the three models. As for the tensor train model, it is even worse than the linear regression model. We further illustrate the recovered error of $\mathbf{W}$  with different levels of noises in Figure~\ref{fig:sigma3}. It demonstrates that the weight recovered by the tensor ring model has the best tolerance with respect to various injected noises.

% Following should be deleted.
% We analyze the TR model in this part. Firstly, we generate $x$ from a Gaussian distribution ${{N}(0, 0.5)}$. Then calculate $y$ with a distribution of $\mathbf{y} = \mathbf{W}\mathbf{x}$. Finally, we add Gaussian noises with different variances into ${x}$ to build datasets. 
% % The variance of noises can reveal the degree of pollution in training set. 
% As shown in Figure~\ref{fig:truth3}, we construct a low-rank W-truth with a shape of $81 \times 81$. We set all ranks as 3, input shape as ${3 \times 3 \times 3 \times 3}$, and output shape as ${3 \times 3 \times 3 \times 3}$ in both TR and TT model. This setting can keep the quantities of the parameters in each model equal. We use root-mean-square error(RMSE) to evaluate the performances of these models. Figure~\ref{fig:sigma3} shows the results of the three models with noises of different variances. Intuitionally, the proposed TR model always performs the best among the three models. As for TT model, it is even worse than the linear regression model. And the recovery result of Ground truth $\mathbf{W}$, when the variance of the noise is 0.05, are illustrated in  Figure~\ref{fig:model-analysis}.

% Figures of Sensitivity Analysis

\subsection{Experiments on the UCF11 Dataset}
The UCF11 dataset contains 1600 video clips of a resolution ${320 \times 240}$ divided into 11 action categories (e.g., basketball shooting, biking/cycling, diving, etc.). Each category consist of 25 groups of video, within more than 4 clips in one group. It is a challenging dataset due to large variations in camera motion, object appearance and pose, object scale, cluttered background, and so on.

In this part, we conduct two experiments described as ``End-to-End Training'' and ``Pre-train with CNN'' on this dataset. In the ``End-to-End Training'', we compare the proposed TR-LSTM model with other decomposition models (eg. TT-LSTM~\cite{DBLP:conf/icml/YangKT17} and BT-LSTM~\cite{Ye_2018_CVPR}) to show the superior performance. And in another experiment, we apply decomposition on a more general model, achieving a better performance with less parameters.

% Figures of "with frames"
\begin{figure}[ht]
\centering
\subfigure[Compression Ratio]{
    \centering
    \label{fig:with-frames-compress-ratio}
    \includegraphics[scale=0.3]{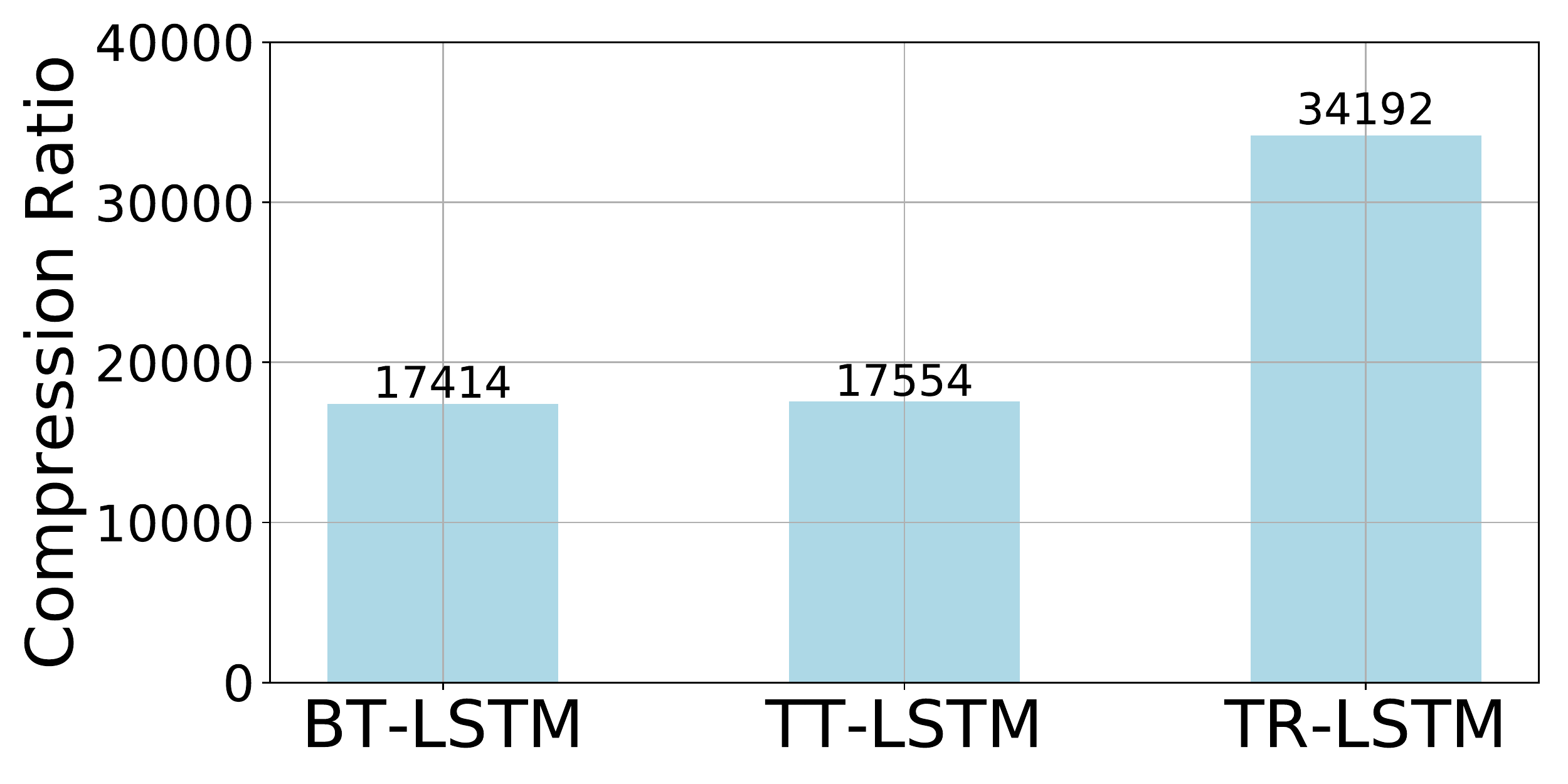}
}
\subfigure[Train Loss]{
    \centering
    \label{fig:with-frames-train-loss}
    \includegraphics[scale=0.3]{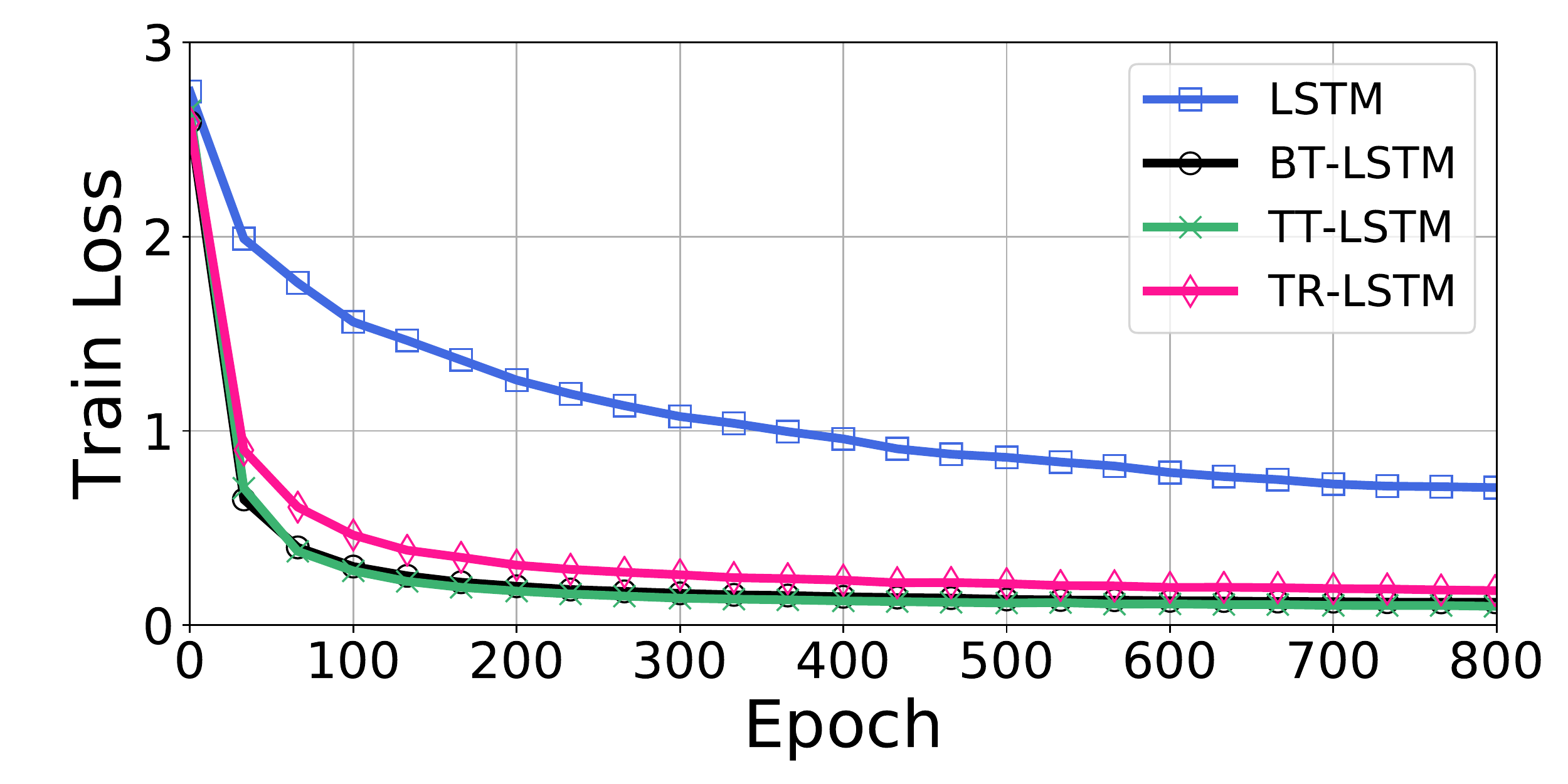}
}
\subfigure[Test Accuracy]{
    \centering
    \label{fig:with-frames-val-accu}
    \includegraphics[scale=0.3]{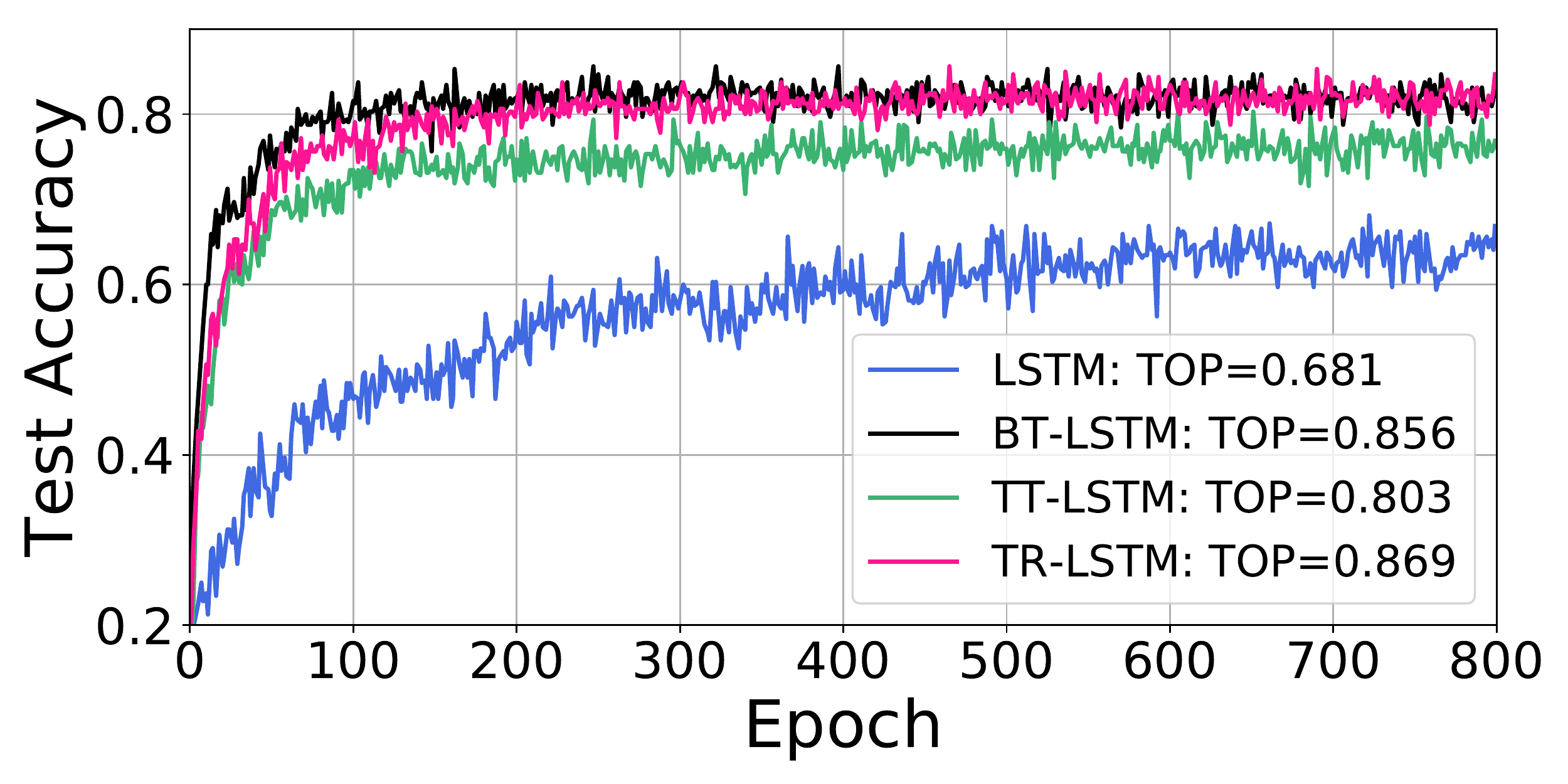}
}
\caption{The results of ``End-to-End Training'' on UCF11 dataset. (a) shows the different compression ratio based on the vanilla LSTM. (b) and (c) shows the training and testing details.}
\label{fig:with-frames}
\end{figure}

\paragraph{End-to-End Training}
Recent years, some tensor decomposition models are proposed to classify videos like TT-LSTM~\cite{DBLP:conf/icml/YangKT17}, BT-LSTM~\cite{Ye_2018_CVPR} and others. For the reason that they use the end-to-end model for training, we set this experiment to compare with them. In this experiments, we scale down the original resolution to ${160 \times 120}$, and sample 6 frames from each video clip randomly as the input data. Since every frame is RGB, the input data vector at each step is ${160 \times 120 \times 3 = 57600}$, and there are 6 steps in every sample. We set the hidden layer as 256. So there should be a fully-connected layer of ${4 \times 57600 \times 256 = 58982400}$ parameters to achieve the mapping for the standard LSTM.

\begin{table}[h]
% \small
\centering
\caption{Results of ``End-to-End Training'' on UCF11 reported in literature. TT-LSTM was reported in \cite{DBLP:conf/icml/YangKT17} while the BT-LSTM was reported in \cite{Ye_2018_CVPR}.}
\begin{tabular}{c|c|c}
\hline
Method& \#Params& Accuracy\\
\hline\hline
LSTM& 59M& {0.697}\\
TT-LSTM & 3360& 0.796\\
BT-LSTM& 3387& 0.853\\
TR-LSTM& 1725& \textbf{0.869}\\
\hline
\end{tabular}

\vspace{-2ex}
\label{tab:with-frames}
\end{table}

We compare our model with BT-LSTM and TT-LSTM, while using a standard LSTM as a baseline. The hyper-parameters in BT-LSTM and TT-LSTM are set as announced in their papers. Figure~\ref{fig:with-frames-val-accu} shows all decomposition methods converging faster than the LSTM. The accuracy of BT-LSTM is 0.856 which is much higher than TT-LSTM with 0.803 while the LSTM only gain an accuracy of 0.681. In our TR-LSTM, the shape of input tensor is ${4 \times 2 \times 5 \times 8 \times 6 \times 5 \times 3 \times 2}$, the output tensor's shape is set as ${4 \times 4 \times 2 \times 4 \times 2}$ and all the TR-ranks are set as 5 except $R_0 = R_d = 10$. %except one of them is 10. 
Results are compared in Table~\ref{tab:with-frames}. With 1725 parameters in our model, about half of TT-LSTM and BT-LSTM with parameters 3360 and 3387 respectively. We gain the top accuracy 0.869, showing the outstanding performance of our model in this experiment.

\begin{table}[h]
\centering
\caption{The state-of-the-art performance on UCF11.}
\begin{small}
\begin{tabular}{l|c}
\hline
Method& Accuracy\\
\hline
\hline
\cite{DBLP:conf/cvpr/HasanR14}& {54.5\%}\\
% \hline
\cite{DBLP:conf/cvpr/LiuLS09}& {71.2\%}\\
% \hline
\cite{DBLP:conf/eccv/Ikizler-CinbisS10}& {75.2\%}\\
% \hline
\cite{liu2013spatial}& {76.1\%}\\
% \hline
\cite{DBLP:journals/corr/SharmaKS15}& {85.0\%}\\
% \hline
\cite{DBLP:conf/cvpr/WangKSL11}& {84.2\%}\\
% \hline
\cite{DBLP:journals/corr/SharmaKS15}& {84.9\%}\\
% \hline
\cite{DBLP:journals/pr/ChoLCO14}& {88.0\%}\\
% \hline
% \cite{DBLP:journals/corr/RavanbakhshMRMD15}& {89.5\%}\\
% \hline
\cite{DBLP:conf/wacv/GammulleDSF17}& \textbf{94.6\%}\\
\hline
\hline
CNN + LSTM& {92.3\%}\\
% \hline
CNN + TR-LSTM& {93.8}\%\\
\hline
\end{tabular}
\end{small}

\label{tab:with-features-compare}

\end{table}

\paragraph{Pre-train with CNN}
Recently, some methods based on RNNs achieved higher accuracy by using the extracted feature as input vectors in computer vision~\cite{DBLP:conf/cvpr/DonahueHGRVDS15}. Compared with using frames as input data, extracted features are more compact. But there is still some room for improving the ability of the models. The over-parametric problem is just partial solved. To get better performance, we use extracted features via the CNN model Inception-V3 as input data to LSTM.

We set the size of the hidden layer as ${32 \times 64 = 2048}$, which is consistent with the size of the output via Inception-V3. After using the extracted feature as the inputs of LSTM, the accuracy of the vanilla LSTM attains 0.923. At the same time, the accuracy of our TR-LSTM model whose ranks are set as ${40 \times 60 \times 48 \times 48}$ achieves 93.8. By replacing the standard LSTM with our model, a compression ratio of 25 can be obtained.
% and our TR-LSTM model are faster than the standard LSTM to complete one training epoch.
We compare some state-of-the-art methods
% (e.g., Visual Attention~\cite{DBLP:journals/corr/SharmaKS15}, Two Stream LSTM\cite{DBLP:conf/wacv/GammulleDSF17}, etc.) 
in Table~\ref{tab:with-features-compare} on UCF11. The Two Stream LSTM\cite{DBLP:conf/wacv/GammulleDSF17} with highest accuracy has more than 141M parameters. The TR-LSTM can be used to replace the vanilla LSTMs in the Two Stream LSTM model to reduce the parameters.

% % Figures of "with features"
% \begin{figure*}[h]
% \subfigure[Compress Ratio]{
%     \centering
%     \label{fig:with-features-compress-ratio}
%     \includegraphics[scale=0.2]{figure/feature11_parameters.pdf}
% }
% \hfill
% \subfigure[Train Loss]{
%     \centering
%     \label{fig:with-features-train-loss}
%     \includegraphics[scale=0.2]{figure/feature11_train_loss.pdf}
% }
% \hfill
% \subfigure[Test Accuracy]{
%     \centering
%     \label{fig:with-features-val-accu}
%     \includegraphics[scale=0.2]{figure/feature11_val_acc.pdf}
% }
% \caption{Based on CNN Training with hidden layer of 256 on UCF11}
% \label{fig:with-features}
% \end{figure*}

% \begin{figure}[]
% \centering

% \subfigure[Val Loss]{
%     \centering
%     \label{fig:hmdb51-train-loss}
%     \includegraphics[scale=0.6]{figure/hmdb51_val_loss.png}
% }

% \subfigure[Val Accuracy]{
%     \centering
%     \label{fig:hmdb51-val-accu}
%     \includegraphics[scale=0.6]{figure/hmdb51_val_acc.png}
% }

% \caption{HMDB51 Results}
% \label{fig:hmdb51-results}
% \end{figure}

\subsection{Experiments on the HMDB51 Dataset}
The HMDB51 dataset is a large collection of realistic videos from various sources, such as movies and web videos. The dataset is composed of 6766 video clips from 51 action categories.

% \begin{table}[h]
% \centering

% \begin{tabular}{|c|c|c|}
% \hline
% Method& Params & Accuracy\\
% \hline
% \hline
% CNN + LSTM-512& 2097152& {57.3\%}\\
% \hline
% CNN + TR-MPS-LSTM-512& 234496& {60.6\%}\\
% \hline
% \hline
% CNN + LSTM-2048& 16777216& {62.9\%}\\
% \hline
% CNN + TR-MPS-LSTM-2048& 663552& \textbf{63.8}\%\\
% \hline
% \end{tabular}

% \caption{Results on HMDB51}
% \label{tab:hmdb51-compare}
% \end{table}

\begin{table}[h]
\centering
\caption{Comparison with state-of-the-art results on HMDB51. The best accuracy is 0.664 from the I3D model reported in \cite{DBLP:conf/cvpr/CarreiraZ17}, which bases on 3D ConvNets and is not RNN-based method.}
\begin{small}
\begin{tabular}{l|c}
\hline
Method& Accuracy\\
\hline
\hline
\cite{DBLP:conf/cvpr/CaiWPQ14}& {55.9\%}\\
% \hline
\cite{DBLP:conf/iccv/WangS13a}& {57.2\%}\\
% \hline
\cite{DBLP:conf/iccv/TranBFTP15}& {56.8\%}\\
% \hline
\cite{DBLP:conf/cvpr/FeichtenhoferPZ16}& {56.8\%}\\
% \hline
\cite{DBLP:conf/cvpr/Wang0T15}& {63.2\%}\\
% \hline
\cite{DBLP:conf/cvpr/CarreiraZ17}& {\textbf{66.4\%}}\\
% \hline
\cite{DBLP:conf/cvpr/NiMYY15}& {65.5\%}\\
% \hline\cite{DBLP:conf/eccv/JiangDXLN12}& {40.7\%}\\
% \hline
\cite{DBLP:conf/cvpr/JainJB13}& {52.1\%}\\
% \hline
\cite{DBLP:conf/iccv/ZhuWYZT13}& {54.0\%}\\
% \hline
% \cite{DBLP:conf/eccv/PengZQP14}& {66.8\%}\\
\hline
\hline
CNN + LSTM & {62.9\%}\\
% \hline
CNN + TR-LSTM & {63.8}\%\\
\hline
\end{tabular}
\end{small}

\label{tab:hmdb51-bests}
\end{table}

% Why use two size? To demonstrate improving the size is reasonable.
In this experiment, we still use extracted features as the input vector via Inception-V3 and reshape it into ${64 \times 32}$. We sample 12 frames from each video clip randomly and be processed through the CNN as the input data. The shape of hidden layer tensor is set as ${32 \times 64 = 2048}$. The ranks of our TR-LSTM are ${40 \times 60 \times 48 \times 48}$. Some of the state-of-the-art models like I3D~\cite{DBLP:conf/cvpr/CarreiraZ17} are presented in Table~\ref{tab:hmdb51-bests}. The I3D model with highest accuracy based on 3D ConvNets, which is not RNN-based method, has 25M parameters, while the TR-LSTM model only has 0.7M parameters. The TR-LSTM gains a higher accuracy of 63.8\% than the standard LSTM with a compressing ratio of 25.

\section{Related Work}

In the past decades,  a number of variants of recurrent neural networks (RNNs) were proposed to capture sequential information more accurately~\cite{LiuHBDBX18}. However, when dealing with the large input data, in the field of computer vision, the input-to-hidden weight matrix owns loads of parameters. The limitation of computation resources and the severe over-fitting problem are emerging. Some methods used CNNs as feature extractors to pre-processing the input data into a more compact way~\cite{DBLP:conf/cvpr/DonahueHGRVDS15}. These methods have improved the classification accuracy, but the over-parametric problem is just still partially solved.
% Distillation is an effective method to obtain compact models by distilling knowledge from specialist models~\cite{DBLP:journals/corr/HintonVD15}, but lacking of effective implementation in RNNs~\cite{???}. 
In this work, we focus on designing low-rank structure to replace the redundant input-to-hidden weight matrix in RNNs, while compressing the whole model and maintaining the model performance.

The most straight-forward way to apply low-rank constraint is implementing matrix decomposition on weight matrices. Singular Value Decomposition (SVD) has been 
% used in decomposing weight matrices in feed-forward neural networks for the large vocabulary continuous speech recognition (LVCSR) task~\cite{DBLP:conf/interspeech/XueLG13}, achieving $3\times$ in compression ratio with a loss in accuracy. SVD was also 
applied in convolutional neural networks to reduce parameters~\cite{DBLP:conf/nips/DentonZBLF14} but incurred a loss in model performance. Besides, the compression ratio is limited because of the rank in matrix decomposition still relatively large. 

Compared with matrix decomposition,  tensor decomposition~\cite{XuYQ15,ZheZWLXQG16,HaoLYX18,He2018,Liu2018} conducts data decomposition in a higher dimension, capturing higher-order correlations while maintaining several orders of fewer parameters~\cite{5_tucker_FC_CONV,DBLP:conf/nips/NovikovPOV15}. Among these methods, \cite{DBLP:journals/corr/LebedevGROL14} utilized CP decomposition to speed up convolution computation, which has the similar design philosophy with the widely-used depth-wise separable convolutions~\cite{DBLP:journals/corr/HowardZCKWWAA17}. However, the instability issue~\cite{DBLP:journals/siammax/SilvaL08} hinders the low-rank CP decomposition from solving many important computer vision tasks. \cite{5_tucker_FC_CONV} used Tucker decomposition to decompose both the convolution layer and the fully connected layer, reducing run-time and energy significantly in mobile applications with minor accuracy drop. 
% Despite the success in applying CP and Tucker decomposition in neural networks, the instability of CP and the exponential increase of parameters in widening networks of Tucker will hinder them from building more efficient models. 
Block-Term tensor decomposition combines the CP and Tucker by summing up multiple Tucker models to overcome their drawbacks and has obtained a better performance in RNNs~\cite{Ye_2018_CVPR}. However, the computation of the core tensor in the Tucker model is highly inefficient due to the complex tensor flatten and permutation operations. Recent years, the tensor train decomposition also used to substitute the redundant fully connected layer in both CNNs and RNNs~\cite{DBLP:conf/nips/NovikovPOV15,DBLP:conf/icml/YangKT17}, preserving the performance while reducing the number of parameters significantly up to 40 times.

% \jmc{  The computation in tensor train is more efficient but suffers from difficulty in finding optimal rank settings~\cite{DBLP:journals/corr/ZhaoZXZC16}, losing in model performance.}

But tensor train decomposition has some limitations: 1) certain constraints for TT-rank, i.e., the ranks of the first and last factors are restricted to be 1, limiting its representation power and flexibility. %This may result in not being found the best representation for a particular data tensor.
2) A strict order must be followed when multiplying TT cores, so that the alignment of the tensor dimensions is extremely important in obtaining the optimized TT cores, but it is still a challenging issue in finding the best alignment. 
In this paper, we use the Tensor Ring(TR) decomposition~\cite{DBLP:journals/corr/ZhaoZXZC16} to overcome the drawbacks in TTD,  while achieving more computation efficiency than BT decomposition.
% TR can be considered as a linear combination of tensor train models but constructed in a circular structure. The circular multi-linear tensor products result in permutation invariant, leading to a more flexible ranks' setting. 
% ~\cite{Wang_2018_CVPR} used TR decomposition to complete FC and conv layer compression , which achieved the compression of $243\times$ on the Wide ResNet model with only losing 2.3\% accuracy .

%\section{Discuss}

\section{Conclusion}
In this paper, we applied TRD to plain RNNs to replace the over-parametric input-to-hidden weight matrix when dealing with high-dimensional input data. The low-rank structure of TRD can capture the correlation between feature dimensions with fewer orders of magnitude parameters. Our TR-LSTM model achieved best compression ratio with the highest classification accuracy on UCF11 dataset among other end-to-end training RNNs based on low-rank methods. At the same time, when processing the extracted feature through InceptionV3 as the input vector, our TR-LSTM model can still compress the LSTM while improving the accuracy. We believe that our models provide fundamental modules for RNNs, and can be widely used to handle large input data. In future work, since our models are easy to be extended, we want to apply our models to more advanced RNN structures~\cite{DBLP:conf/wacv/GammulleDSF17} to get better performance.

\section{Acknowledgments}

We thank the anonymous reviewers for valuable comments to improve the quality of our paper. This work was partially supported by National Natural Science Foundation of China (Nos.61572111 and 61876034), and a Fundamental Research Fund for the Central Universities of China (No.ZYGX2016Z003).

%\section{Future Work}

\bibliographystyle{aaai}
\bibliography{ref}

\clearpage

\section{Appendix}
\subsection{Complexity Analysis}
\label{complexity-ana}

\subsubsection{Complexity in Forward Process}
Since the weight tensor has been decomposed into the form of TRD with $n+m$ core tensors, the order of multiplication among input tensor and core tensors in Equation~(\ref{eq:mps_net}) determines the computational cost. In our implementation, we multiply the input tensor with input core tensors and output core tensors sequentially. We can rewrite the Equation~(\ref{eq:mps_net}) as:
\begin{align}
\pmb{\mathcal{Y}}=&{\pmb{\mathcal{X}}}{\times^{1}_{2}}{\pmb{\mathcal{G}}^{(1)}}{\times^{1, n+1}_{2, 1}}{\pmb{\mathcal{G}}^{(2)}}{\dots}{\times^{1, n-N+3}_{2, 1}}{\pmb{\mathcal{G}}^{(N)}}{\dots}\notag \\
&{\times^{1, 3}_{2, 1}}{\pmb{\mathcal{G}}^{(n)}}{\times^{2}_{1}}{\pmb{\mathcal{G}}^{(n+1)}}
% {\times^{3}_{1}}{\pmb{\mathcal{G}}^{(n+2)}}
{\dots}{\times^{M+1}_{1}}{\pmb{\mathcal{G}}^{(n+M)}}{\dots}\notag \\
&{\times^{m}_{1}}{\pmb{\mathcal{G}}^{(n+m-1)}}{\times^{1,m+1}_{3,1}}{\pmb{\mathcal{G}}^{(n+m)}}
\label{eq:complexity}
\end{align}

In Equation~(\ref{eq:complexity}), the symbols $\times^{a}_{b}$ and $\times^{a, b, c{\dots}}_{{\alpha}, {\beta}, {\theta}{\dots}}$ denote the Tensor Contraction Operation as described in~\cite{DBLP:journals/corr/CichockiLOPZM16}, which is a fundamental and important operation in tensor networks.  Tensor contraction can be considered as a higher-dimensional analogue of matrix multiplication, inner product, and outer product. Note the $N$-th component can be formulated as ${\times^{1, n-N+3}_{2, 1}}{\pmb{\mathcal{G}}^{(N)}}, N \in \{2,3,\dots,n\}$ and the $(n+M)$-th component can be formulated as ${\times^{M+1}_{1}}{\pmb{\mathcal{G}}^{(n+M)}}, M \in \{1,2,\dots,m-1\}$.

Our model is trained via Back Propagation Through Time. In Equation~(\ref{eq:complexity}), according to the left-to-right multiplication order, our forward computational complexity reaches ${\pmb{\mathcal{O}}(nIR^3+mOR^3)}$ while the forward space complexity is ${\pmb{\mathcal{O}}(IR^2)}$, where all the ranks in our model are set to $R$.

\begin{table}[h]
% \small
\centering
\caption{Comparison among vanilla RNN, TT-RNN~\cite{DBLP:conf/icml/YangKT17}, BT-RNN~\cite{Ye_2018_CVPR} and our model TR-RNN on complexity and memory usage. TT-RNN, BT-RNN and TR-RNN are all set in same rank $R$. Here, $d$ denotes the number of factors in BT-RNN and the number of cores in TT-RNN. $O_{max}=max_k(O_k)$ and $I_{max}=max_k(I_k), k \in \{ 1, 2, \dots, d \}$. $\Gamma$ represents $max\{I, {I_{max}}O\}$. Notaions  $F$ and $B$ denote forward process and backward process individually.}

\begin{tabular}{l|l|l}
\hline
Method& Time& Memory\\
\hline\hline
RNN $F$& ${\pmb{\mathcal{O}}(IO)}$& ${\pmb{\mathcal{O}}(IO)}$\\
RNN $B$& ${\pmb{\mathcal{O}}(IO)}$& ${\pmb{\mathcal{O}}(IO)}$\\
TT-RNN $F$& ${\pmb{\mathcal{O}}(dI{R^2}{O_{max}})}$& ${\pmb{\mathcal{O}}(RI)}$\\
TT-RNN $B$& ${\pmb{\mathcal{O}}({d^2}I{R^4}{O_{max}})}$& ${\pmb{\mathcal{O}}({R^3}I)}$\\
BT-RNN $F$& ${\pmb{\mathcal{O}}(NdI{R^d}{O_{max}})}$& ${\pmb{\mathcal{O}}({R^d}I)}$\\
BT-RNN $B$& ${\pmb{\mathcal{O}}(N{d^2}I{R^d}{O_{max}})}$& ${\pmb{\mathcal{O}}({R^d}I)}$\\
TR-RNN $F$& ${\pmb{\mathcal{O}}(nIR^3+mOR^3)}$& ${\pmb{\mathcal{O}}(R^2I)}$\\
TR-RNN $B$& ${\pmb{\mathcal{O}}(ndIR^5+md{I_{max}}OR^5)}$& ${\pmb{\mathcal{O}}({R^4}{\Gamma})}$\\
\hline
\end{tabular}

\vspace{-2ex}
\label{tab:compare-complexity}
\end{table}

\subsubsection{Complexity in Backward Process}
We implement our back propagation with considering the backward complexity analyze of TT~\cite{DBLP:conf/nips/NovikovPOV15}. The backward computational complexity of different cores is different due to the multiplication order, so we choose $\frac{\partial \pmb{\mathcal{L}}}{\partial \pmb{\mathcal{G}}^{(k)}}$, $k \in \{2,3,\dots,n-2\}$ to represent the backward complexity of each core, which has the highest computational and space complexity in all the cores:
\begin{align}
\frac{\partial \pmb{\mathcal{L}}}{\partial \pmb{\mathcal{G}}^{(k)}}=&{\pmb{\mathcal{X}}}{\times^{1}_{2}}{\pmb{\mathcal{G}}^{(1)}}{\times^{1, n+1}_{2, 1}}{\pmb{\mathcal{G}}^{(2)}}{\dots}{\times^{1, n-{N_1}+3}_{2,1}}{\pmb{\mathcal{G}}^{({N_1})}}{\dots}\notag \\
&{\times^{1, n-k+4}_{2, 1}}{\pmb{\mathcal{G}}^{(k-1)}}{\times^{2}_{2}}{\pmb{\mathcal{G}}^{(k+1)}}{\times^{2, n-k+4}_{2, 1}}{\pmb{\mathcal{G}}^{(k+2)}}{\dots}\notag \\
&{\times^{2, n-k-{N_2}+6}_{2, 1}}{\pmb{\mathcal{G}}^{(k+{N_2})}}{\dots}{\times^{2, 6}_{2, 1}}{\pmb{\mathcal{G}}^{(n)}}\notag \\
&{\times^{5}_{1}}{\pmb{\mathcal{G}}^{(n+1)}}{\dots}{\times^{4+M}_{1}}{\pmb{\mathcal{G}}^{(n+M)}}{\dots}{\times^{m+3}_{1}}{\pmb{\mathcal{G}}^{(n+m-1)}}\notag \\
&{\times^{2, m+4}_{3, 1}}{\pmb{\mathcal{G}}^{(n+m)}}{\times^{4, 5{\dots}, m+3}_{1, 2{\dots}, m}}{\frac{\partial \pmb{\mathcal{L}}}{\partial \pmb{\mathcal{Y}}}}
\label{eq:complexity-back}
\end{align}

In Equation~(\ref{eq:complexity-back}), the notations $N_1$, $N_2$ and $M$ are the same as $N$ and $M$ in Equation~(\ref{eq:complexity}) and here  $N_1 \in \{2,3,\dots,k-1\}$, $N_2 \in \{k+2,k+3,\dots,n\}$, $M \in \{1,2,\dots,m-1\}$. The backward computational complexity is ${\pmb{\mathcal{O}}(nIR^5+m{I_{max}}OR^5)}$. Note that the number of cores is $d$ in our model, the total backward computational complexity reaches to ${\pmb{\mathcal{O}}(ndIR^5+md{I_{max}}OR^5)}$ while the backward space complexity is ${\pmb{\mathcal{O}}({R^4}max\{I, {I_{max}}O\})}$. The statistics of comparison with some other compressing methods are shown in Table~\ref{tab:compare-complexity}.

\end{document}